\documentclass[10pt,twocolumn,letterpaper]{article}

\usepackage{cvpr}
\usepackage{times}
\usepackage{epsfig}
\usepackage{graphicx}
\usepackage{amsmath}
\usepackage{amssymb}
\usepackage{subfigure}
\usepackage{url}
\usepackage[british,UKenglish,USenglish,english,american]{babel}

\def\relus{{\ensuremath{\operatorname{\mathop{ReLU6}\,}}}}
\def\depthwise{{\ensuremath{\operatorname{\mathop{dwise}\,}}}}

\newcommand{\ve}[1]{\mathbf{#1}} 

\newcommand{\fp}[2]{\frac{\partial{#1}}{\partial{#2}}} 
\newcommand{\dvp}[2]{\frac{\partial}{\partial{#2}}#1} 

\usepackage[breaklinks=true,bookmarks=false]{hyperref}

\cvprfinalcopy 

\def\cvprPaperID{****} 

\begin{document}

\title{SeesawNet: Convolution Neural Network With Uneven Group Convolution}

\author{Jintao Zhang\\
DiDiChuxing\\
{\tt\small jtzhangcas@gmail.com}
}

\maketitle

\begin{abstract}
In this paper, we are interested in boosting the representation capability of convolution neural networks which utilizing the inverted residual structure. Based on the success of Inverted Residual structure(Sandler et al. 2018) and Interleaved Low-Rank Group Convolutions(Sun et al. 2018), we rethink this two pattern of neural network structure, rather than NAS(Neural architecture search) method(Zoph and Le. 2017; Pham et al. 2018; Liu et al. 2018b), we introduce uneven point-wise group convolution, which provide a novel search space for designing basic blocks to obtain better trade-off between representation capability and computational cost. Meanwhile, we propose two novel information flow patterns that will enable cross-group information flow for multiple group convolution layers with and without any channel permute/shuffle operation. Dense experiments on image classification task show that our proposed model, named Seesaw-Net, achieves state-of-the-art(SOTA) performance with limited computation and memory cost. Our code will be open-source and available together with pre-trained models.
\end{abstract}

\section{Introduction}
Convolutional neural networks (CNNs), inspired by biological neural networks(BNNs), comes to be leading architecture in deep learning that are used for computer vision as well as many other Techniques, since the milestone work of AlexNet(Alex Krizhevsky et al. 2012) and all the other classical models(Simonyan et al. 2014; Szegedy et al. 2015; He et al. 2015) signicantly improved the SOTA in vision related tasks(Russakovsky et al. 2015). However, deeper and larger modern models comsuming much more hardware resources and time may not always be a wise choice for real-time application (RTA) or when applications including sensitive customer data of which the privacy and security need protection, especially on resource-constrained platforms, such as mobile and wearable devices. Besides, power consumption is also a critical but always neglected issue especially for implementation of Always-On features.

Given restricted computational resources and bandwidth available on mobile devices, most of recent research has focused on designing mobile CNN models by reducing the depth/width of the network and replace traditional expensive operations with lightweight ones, such as depthwise convolution (Howard et al. 2017) and group convolution (Zhang et al. 2018). Generally, designing a resource-constrained mobile model is difficult, designing a scalable model which could achieve comparable performance whenever scaled up or down compared to corresponding SOTA work is even more challenging task. What makes things more complicated is that, according to our years of experience and related work of others[Ignatov et al. 2018], large variation will occur as a result of different optimization status on diverse training or inference platforms.

In this paper, we introduce uneven point-wise group convolution for neural network blocks design,  as well as two novel information flow patterns that enable cross-group information flow for multiple group convolution layers with and without any channel permute/shuffle operation, which utilize sparse grouping kernels to compose a robust neural block. We start from Mobilenet-V2 (Sandler et al. 2018) which is composed of Inverted Residuals and Linear Bottlenecks as well as IGCV3(Sun et al. 2018) which combines low-rank and sparse kernels for network design.

Our main contribution is introducing a novel space for NAS(Neural Architecture Search): utilizing hyber grouping strategies to obtain structure-sparse neural networks with sufficient representation capacity. We expected the classical average grouping convolution is just the most simple variant of group convoution, that is, standard convolution filters are artificially averaged into pre-defined number of groups. To the best of our knowledge, the classical average grouping convolution is an artificial and technical tricks, whose variant grouping strategy is rarely mentioned in previous work on efficient network design, although condensenet introduce Learned Group Convolution but the output features of each block still use even grouping.

\section{Related Work}
\subsection{Efficient architecture design}
Improving the resource efficiency of CNN models has been an active research issue during last several years. Mainstream of engineering approaches include:
1) Low-Bit quantization the weights and/or activations of a baseline CNN model (Han, Mao, and Dally 2015; Jacob et al. 2018), or 2) reducing computational cost by pruning less important filters (Gordon et al. 2018; Yang et al. 2018) based on value of the weights and bias during or after training.

Another common approach is efficient neural architectures design, which we note as our main design method. SqueezeNet(Iandola et al. 2016) reduces the amount of parameters and
computational cost by pervasively using lower-cost 1x1 convolutions and reducing filter sizes; MobileNet (Howard et al. 2017) extensively employs depthwise separable convolution to minimize computation density; ShuffleNet (Zhang et al. 2018) utilizes low-cost pointwise group convolution and introduce channel shuffle operation; Condensenet (Huang et al. 2018) learns to connect group convolutions across layers; Recently, Mobilenet-V2 (Sandler et al. 2018) and  IGCV3(Sun et al. 2018) achieved state-of-the-art results on general vision tasks among mobile-size models with resource-efficient inverted residuals and linear bottleneck structure.
\subsection{Neural architecture search}
Neural Architecture Search (NAS), the process of automating architecture engineering, is a logical next step in automating machine learning. Recent research commonly utilize NAS to optimize convolutional architectures on certain task of interest. However, applying NAS, or any other search methods, directly to a complicated task on large dataset, such as the ImageNet dataset, is too computationally expensive.

Search space, the core dimension of NAS, defines which architectures can be represented in principle. However, according previous work[Elsken et al. 2018], the key limination of NAS is clear that NAS will introduces human bias, which may prevent finding novel architectural building blocks that go beyond he current human knowledge, since search space is usually predefined according to only known structures.
\section{Several questions}
\subsection{Group Convolution}
Group convolution was firstly introduced in the seminal AlexNet paper in 2012. As explained by the authors, their primary motivation was to allow the training of the network over two GPUs. However, it has been well demonstrated its effectiveness in ResNeXt architecture. Depthwise seperate convolution decouple spatial and channel feature, proved to be successful by series of efficient network architecture. Recently, MobileNet-v1/v2 and Shufflenet-v1/v2 utilizes group convolution and depthwise separable convolutions to gains state-of-the-art results among lightweight models. However, to the best of our knowledge, only average grouping has been applied in neural network architecture design ever. Our work generalizes group convolution in a novel pattern-seesaw group convoution.

\subsection{Channel Shuffle/Permute Operation}
For most deep learning framework users, channel shuffle/permute operations will introduce extra cost, especially for IGCV3 which include 2 channel permute/shuffle operation per basic network block, compared with Shufflenet-v1/v2 which include 1 channel shuffle/permutation operation per basic network block. Meanwhile, for underlying programming language users, the time consumption resulted from channel shuffle/permute operations could be avoided or at least reduced for typical network architectures.
Our seesawnet include only one shuffle/permute operation for all building blocks including the residual blocks, which breaks the original rules introduced by the residual networks but achieve SOTA performance on dense experiments. We expect this will save certain cost for deep learning framework users.

\subsection{Sparse Kernels}
Group convolution is a pre-defined structured-sparse matrix, which is widely used in the mobile models. According to many classical works, sparse kernel is prefered to build robust model. However, average grouping might not be the ground truth, since no evidence show that average grouping neural units exist in BNNs. Furthermore, we notice that recent efficient neural networks which utilize group convolution could achieve comparable or better performance than other model.

\subsection{Layer diversity}
MnasNet has notable better accuracy-latency trade-offs over those variants, suggesting the importance of layer diversity in resource constrained CNN models. Compared to Mobilenet-v2 We can notice that key super parameters like repeat time of building blocks changed greatly when 5x5 convolution kernel was used. However, instead of NAS with alternative super parameters like kernel size and repeat time, we propose a novel grouping convolution pattern so as to add diversity to the layers or build blocks of neural network design.

\subsection{Metric of computation complexity}
According to series of previous work(MobileNet-v1/Shufflenet/MobileNet-v2...), flops, rather than the amount of model parameters, should be a reliable guide as the metric of computation complexity. Furthermore, Shufflenet-v2/MnasNet auge that direct metric like speed/fps should be a better direct metric on the target platform. Rather than proposing series of experiment results on specific platforms, we would like to focus on the current fact of Hardware-Software design for deep learning.
Generally, to apply a certain deep learning model(e.g., convolution nueral network model), most researchers and developers would choose deep learning frameworks like CNTK/TensorFlow/Keras/Caffe/Caffe2/Pytorch/Mxnet for online inference and TensorFlow-Lite/ncnn/MACE for offline inference. However, a key bottleneck for these frameworks is that hardware dependency widely exists when performance training/inference different models on different target platforms(Ignatov et al. 2017). We expect this is the reason certain researchers and developers implement specific model with low-level language like C++ or private frameworks for online inference and neon/opencl/cuda based language or private frameworks for offline inference, so as to achieve best performance-efficiency balance compared to deep learning frameworks.

\section{arch design}
\subsection{Seesaw Grouping}
Unlike any group convolution pattern from previous work, we introduce diversity as well as sparsity into neural network design by implement pointwise convolution in Seesaw groups(fig.1). Seesaw grouping, which plays the key role in our design, utilize uneven grouping strategy based on group convolution(we take IGCV3 composed of Interleaved Low-Rank Group Convolutions as our baseline model in this paper).
Classical group convolution networks(IGCV, Shufflenet-v1, IGCV2, IGCV3, Shufflenet-v2) utilize even grouping as a design pattern to add sparsity to neural network models as well as to eliminate resource requirement, Since group convolution replace full connection between all channels with group connection.
IGCV and Shufflenet-v1 proposed the concept of channel permute and channel shuffle in efficient network design, which proved successful in later works(IGCV2, IGCV3, Shufflenet-v2) since channel permute/shuffle operation will help information flow across feature channels.
We introduce uneven grouping strategy to replace shallow even grouping based blocks in IGCV3, since we expect that even grouping is just a special strategy of group convolution which use least parameters and computation amount compared with other grouping strategies.
Seesaw shuffle Grouping block also include channel permute operation which proved helpful in dense experiments.
 
\begin{figure}[htbp]
\centering
\subfigure[Mobilenet-v2]{
\begin{minipage}[t]{0.4\linewidth}
\centering
\includegraphics[width=1.3in]{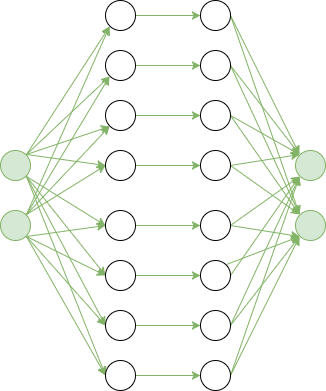}
\end{minipage}%
}%
\subfigure[IGCV3]{
\begin{minipage}[t]{0.4\linewidth}
\centering
\includegraphics[width=1.3in]{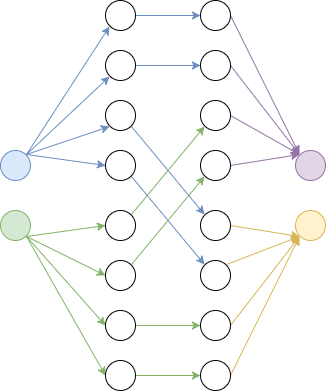}
\end{minipage}%
}%

\subfigure[Seesaw-shuffleNet]{
\begin{minipage}[t]{0.4\linewidth}
\centering
\includegraphics[width=1.3in]{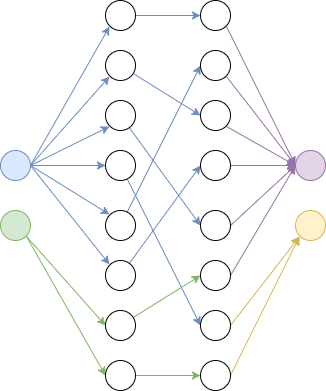}
\end{minipage}
}%
\subfigure[Seesaw-shareNet]{
\begin{minipage}[t]{0.4\linewidth}
\centering
\includegraphics[width=1.3in]{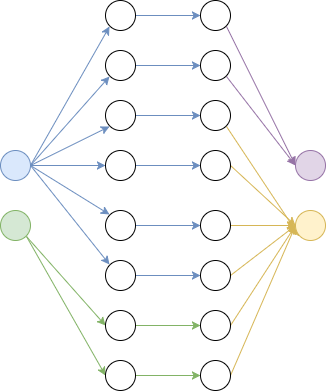}
\end{minipage}
}%
\centering
\caption{ Building blocks list}
\end{figure}

\begin{table}[htbp]
\begin{center}
\begin{tabular}{c|c|c}
\hline
Input & Operator & Output\\
\hline\hline
    $h \times w \times k$ & 1x1 uneven group conv2d & $h \times w \times (tk)$\\
    $h \times w \times tk$& 3x3 \depthwise s=$s$, \relus & $\frac{h}{s} \times \frac{w}{s} \times (tk)$\\ 
    $\frac{h}{s} \times \frac{w}{s} \times tk$ & 1x1 uneven group conv2d & $\frac{h}{s} \times \frac{w}{s} \times k'$\\
\hline
\end{tabular}
\end{center}
\caption{Seesaw block implementation}
\end{table}

\begin{table}[htbp]
\begin{center}
\begin{tabular}{c|c|c|c|c|c}
\hline
Input & Operator & t & c & n & s\\
\hline\hline
$224^2\times3$ &    conv2d                  &  - &  32 & 1 & 2\\
$112^2\times32$ &    uneven-block    &  1 & 16   & 1 & 1 \\
$112^2\times16$ &   uneven-block     &  6 & 24   & 4(2) & 2 \\
$56^2\times24$ &   uneven-block      &  6 & 32   & 6(3) & 2 \\
$28^2\times32$ & uneven-block        &  6 & 64   & 8(4) & 2 \\
$14^2\times64$ &    uneven-block    &  6 & 96   & 6(3) & 1 \\
$14^2\times96$ &    uneven-block     &  6 & 160  & 6(3) & 2 \\
$7^2\times160$ &    even-block     &  6 & 320  & 1 & 1 \\
$7^2\times 320$ & conv2d 1x1       &  - & 1280 & 1 & 1 \\
$7^2\times1280$  & avgpool 7x7     &  - & -    & 1 & - \\   
$1\times1\times 1280$ & conv2d 1x1                      &  - & k    & -& \\

\hline
\end{tabular}
\end{center}
\caption{Hyper parameters of Seesaw-shuffleNet for Imagenet classification task. We adopt the idea of IGCV3 block and remove the relu activation layer after the first linear pointwise group convolution, and double the repeat time of main block to keep same count of non-linearities as mobilenet-V2. Here, repeat time parameters in round brackets stand for 0.5D version which containes only half relu activation layer as mobilenet-V2. For cifar tasks, we keep the same hyper parameters as IGCV3 so as to make fair comparison.}
\end{table}

\subsection{Residual structure}
Compare to MobileNet-v1, a highlight in MobileNet-v2’s building block is the residual connection, with which the information loss is less strong for inverted bottlenecks. IGCV3 adopt the Identity Skip Connections from MobileNet-v2.

 The original Residual Unit in {Deep Residual Learning for Image Recognition} performs the following computation:
\begin{gather}
\ve{y}_{l} = h(\ve{x}_{l}) + \mathcal{F}(\ve{x}_{l}, \mathcal{W}_l), \label{eq:resunit1}\\
\ve{x}_{l+1} = f(\ve{y}_{l}) \label{eq:resunit2}.
\end{gather}

If $f$ is also an identity mapping: $\ve{x}_{l+1} \equiv \ve{y}_{l}$, we can put Eqn.(2) into Eqn.(1) and obtain:
\begin{equation}
\ve{x}_{l+1} = \ve{x}_{l} + \mathcal{F}(\ve{x}_{l}, \mathcal{W}_{l}). \label{eq:additive0}
\end{equation}

Recursively ({\fontsize{8pt}{1em}$\ve{x}_{l+2} = \ve{x}_{l+1} + \mathcal{F}(\ve{x}_{l+1},\mathcal{W}_{l+1})=\ve{x}_{l} + \mathcal{F}(\ve{x}_{l}, \mathcal{W}_{l})+\mathcal{F}(\ve{x}_{l+1}, \mathcal{W}_{l+1})$}, etc.) we will have:
\begin{equation}
\ve{x}_{L} = \ve{x}_{l} + \sum_{i=l}^{L-1}\mathcal{F}(\ve{x}_{i}, \mathcal{W}_{i}), \label{eq:additive}
\end{equation}

Eqn.(4) also leads to nice backward propagation properties.
Denoting the loss function as $\mathcal{E}$, from the chain rule of backpropagation \cite{LeCun1989} we have:
\begin{equation}
\fp{\mathcal{E}}{\ve{x}_{l}}=\fp{\mathcal{E}}{\ve{x}_{L}}\fp{\ve{x}_{L}}{\ve{x}_{l}}=\fp{\mathcal{E}}{\ve{x}_{L}}\left(1+\dvp{\sum_{i=l}^{L-1}\mathcal{F}(\ve{x}_{i}, \mathcal{W}_{i})}{\ve{x}_{l}}\right).\label{eq:grad}
\end{equation}

However, we utilize a special variant of classical residual structure, since we that only one channel permute/shuffle operation included per building block so as to save cost during training and inference. Since channel permute/shuffle operation will mess up original channel order within same layer, then we would like to analysis how the residual structure works in our seesaw unit. In this part, only seesaw-shufflenet is discuessed, since the seesaw-sharenet does not include channel permute/shuffle operation and channel order will keep during propagation.
As a result of seesaw group convolution and channel permute/shuffle operation, 2 situations will occur here:

1. The input channel and the output channel from seesaw building block share the same channel index.

2. The output channel of seesaw building block does not share same channel index as the input channel.

Note that although we remove the second channel permute/shuffle operation from IGCV3, two key features from residual structure and channel shuffler/permute structure are reserved:

1. Eqn.(5) still works for the 2 situations in our seesaw building blocks, since the seesaw grouping strategy will not make any change to the shortcut path.
According to (resnet v1/v2), this will guarantee that the forward and backward signals can be directly propagated from one block to any other block.

2. The channel shuffle/permute operation within our seesaw building blocks will help the information flowing across feature channels, even if our seesaw grouping differs from the even grouping strategy in (shufflenet v1/v2, IGCV series). The importance of cross-group information interchange(Zhang et al. 2018) will be discuessed later.

\section{experiment}

\subsection{Datasets}

{\bfseries CIFAR.} The CIFAR-10 and CIFAR-100 datasets consist of RGB images of size 32*32 pixels, corresponding to 10 and 100 classes, respectively. Both datasets contain 50,000 training images and 10,000 test images from 80 million tiny images. To make a fair comparation with previous work, we use a standard data-augmentation scheme , in which the images are zero-padded with 4 pixels on each side, and then cropped  randomly to produce 32*32 images, then horizontally mirrored with probability 0.5 and normalized by the channel means.

For training, we use the SGD algorithm and train all networks from scratch. We initialize the weights similar to IGCV3, and set the weight decay as 0.0001 and the momentum as 0.9. The initial learning rate is 0.1 and is reduced by a factor 10 at the 200, 300 and 350 training epochs. Each training batch consists of 64 images.

{\bfseries ImageNet.} The ILSVRC 2012 classification dataset contains over 1.2 million images for training and 50,000 images for validation, and all images are labeled from 1000 categories. We adopt the data-augmentation scheme of Inception networks at training time, and perform a rescaling to
256*256 followed by a 224*224 center crop at test time before feeding the input image into the networks, no more extra data-augmentation methods is used.

For training, we use SGD to train the networks with the same hyperparameters (batch size = 96, weight decay = 0.00004 and momentum = 0.9) . Since training from scratch on large dataset like ImageNet is extremely time consuming, unlike the baseline archicture-IGCV3, we train the models with standard data augmentation scheme and learning parameters for 400 epochs from learning rate of 0.045, and scale it by 0.98 every epoch without extra retrain process as IGCV3 which will manually changed the learning rate and data augmentation during training process. We  hold quite similar views like [He et al. 2018] that the number of images, instances, and pixels that have been seen during all training iterations might matters when make comparasion with related works.

\subsection{Comparisons with Other Mobile Networks.}
\begin{table}[!h]
\begin{center}
\begin{tabular}{|l|c|c|c|}
\hline
Model & Params & Multi-Adds & Top1\\
\hline\hline
CondenseNet$^{light}$-94 & 0.33M & 122M & 95.00 \\
IGCV3(0.5D) & 1.4M & 100M & 94.75 \\
Seesaw-shareNet(v0 0.5D) & 1.4M & 104M & 94.99 \\
Seesaw-shareNet(v1 0.5D) & 1.5M & 108M & 94.80 \\
Seesaw-shuffleNet(0.5D) & 1.5M & 108M & \textbf {95.10} \\
\hline
IGCV3(1.0D) & 2.2M & 166M & 95.01 \\
Seesaw-shareNet(v0 1.0D) & 2.3M & 175M & \textbf {95.36} \\
Seesaw-shareNet(v1 1.0D) & 2.4M & 181M & 95.20 \\
Seesaw-shuffleNet(1.0D) & 2.4M & 181M & 95.28 \\
\hline
\end{tabular}
\end{center}
\caption{Comparison of classification Top-1 accuracy (\%) with other mobile networks on the CIFAR-10 dataset.}
\end{table}

\begin{table}[!h]
\begin{center}
\begin{tabular}{|l|c|c|c|}
\hline
Model & Params & Multi-Adds & Top1\\
\hline\hline
CondenseNet$^{light}$-94 & 0.33M & 122M & 75.92 \\
IGCV3(0.5D) & 1.5M & 100M & 77.65 \\
Seesaw-shareNet(v0 0.5D) & 1.5M & 104M & \textbf {77.87} \\
Seesaw-shareNet(v1 0.5D) & 1.6M & 108M & 77.84 \\
Seesaw-shuffleNet(0.5D) & 1.6M & 108M & 77.82 \\
\hline
IGCV3(1.0D) & 2.3M & 166M & 78.24 \\
Seesaw-shareNet(v0 1.0D) & 2.4M & 175M & 78.84 \\
Seesaw-shareNet(v1 1.0D) & 2.5M & 181M & 78.73 \\
Seesaw-shuffleNet(1.0D) & 2.5M & 181M & \textbf {79.26} \\
\hline
\end{tabular}
\end{center}
\caption{Comparison of classification Top-1 accuracy (\%) with other mobile networks on the CIFAR-100 dataset.}
\end{table}

\begin{table}[!h]
\begin{center}
\begin{tabular}{|l|c|c|c|}
\hline
Model & Params & Multi-Adds & Top1\\
\hline\hline
VGG-16 & - & 15,300M & 71.5 \\
ResNet-18 & - & 1,818M & 69.8 \\
Inception V1 & 6.6M & 1,448M & 69.8 \\
NASNet-B (N=4) & 5.3M & 488M & 72.8 \\
NASNet-C (N=3) & 4.9M & 558M & 72.5 \\
MobileNetV1 & 4.2M & 575M & 70.6 \\
ShuffleNet (1.5) & 3.4M & 292M & 71.5 \\
CondenseNet (G=C=8) & 2.9M & 274M & 71.0 \\
MobileNetV2(1.0)  & 3.5M & 314M & 72.0 \\
IGCV3(1.0D) & 3.5M & 318M & 72.2 \\
Seesaw-shuffleNet(1.0D) & 3.6M & 361M & \textbf {72.9} \\

\hline
\end{tabular}
\end{center}
\caption{Comparison of classification Top-1 accuracy (\%) with other networks on the ImageNet dataset.}
\end{table}

\begin{figure}[ht]

\centering
\includegraphics[scale=0.65]{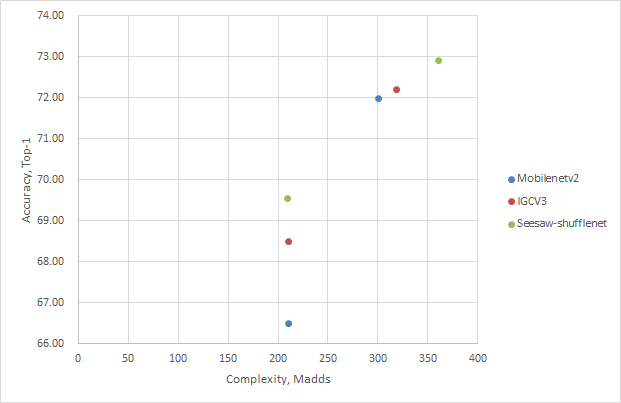}
\caption{Performance Comparison with other SOTA works.( MobileNet V2 data include 1.0 and 0.7W version from(Sandler et al. 2018), IGCV3 data include 1.0 and 0.7D version from (Sun et al. 2018), Seesaw-shufflenet data include 1.0 and 0.5D version)}
\label{fig:label}
\end{figure}

\subsection{Generalization ability across tasks.}
During our experiment, we found the fact that certain part of previous work directly search Neural Architecture on Imagenet for classification tasks. According to those milestone works, we insist on the generalization ability of neural architecture across datasets instead only refer to the performance on single dataset for classification tasks.

\section{Ablation Study}
\subsection{Info flow pattern between grouped channels}
[Zhong et al. 2018] proposed a collection of three shift-based primitives for building efficient compact CNN-based networks, since the classical channel shuffle/permute operation will introduce extra time cost for deep learning framework users. We expected Shufflenet-v2 could eliminate channel shuffle operation because of target-specific optimization and their private implement framework.

To balance efficiency and archicture design, we propose Seesaw-shareNet as a twin network model of Seesaw-shuffleNet. Compared with classical group convolution networks using channel shuffle operation to enable cross-group information flow for multiple group convolution layers, Seesaw-shareNet utilize channel share between adjacent seesaw group convolutions, we expect this will also help cross-group information flow during training and inference. Besides, Compare to Seesaw-shuffleNet, Seesaw-shareNet does not include any channel permute/shuffle operation which will introduce extra time and resource consumption for implemet with mainstream deep learning frameworks and probably will increase implement difficulty at least.

To validate Seesaw-shareNet, we make minimum modification to super parameters(e.g.channels) of Seesaw-shuffleNet so that all channels of point-wise convolutions could be divided into 2 partitions in a ratio of 1:2, no architectural search or benchmark is applied. 

\begin{table}[!h]
\begin{center}
\begin{tabular}{|l|c|c|c|c|}
\hline
Model & permute & Multi-Adds & Top1\\
\hline\hline
CondenseNet-86  &-  & 65M & 95.00 \\
IGCV3(0.5D) & N &  51M & 94.07 \\
IGCV3(0.5D) & Y &  51M & 94.83 \\
Seesaw-shareNet(v1 0.5D) & N & 59M & 94.56 \\
Seesaw-shareNet(v2 0.5D) & Y & 57M & \textbf {95.09} \\
\hline
\end{tabular}
\end{center}
\caption{Comparison of classification Top-1 accuracy (\%) with other tiny mobile networks(with less than 100M flops) on the CIFAR-10 dataset.}
\end{table}

The standalone experiment on cifar-10/cifar-100 validate the power of uneven block convolution strategy. Here, channel permute/shuffle operations are removed to make fair comparation directly with classical even group  convolution.
According to dense experiments, we got 2 conclusions:

(1) Uneven group convolution will help to boost the representation capablilty compared with classical even group convolution. It is easy to prove that classical even group strategy contains least computation cost if input/output channel count and group count is fixed, however, we hold the opinion that it might at least not always the best choice for efficient networks.

(2) Similar to classical channel permute/shuffle, our proposed Seesaw-shuffle block as well as Seesaw-share block, can enable cross-group information flow for multiple group convolution layers. In practical, compared to use the time-consuming channel permute/shuffle operation within common deep learning frameworks, seesaw-share block will help the information flowing across feature channels directly without it.

\subsection{Adaptability on small expansion ratio}
MobileNet-v2 introduce inverted residual structure where the shortcut connections are between the thin bottleneck layers and the intermediate expansion layer uses lightweight depthwise convolutions to filter features as a source of non-linearity. Dense experiments and proof about the expansion ratio are mentioned in (Sandler et al. 2018) to search best configuration for efficiency. However, with limited resources constrain on mobile devices, large expansion ratio will greatly increase computation cost which also will result in lower inference speed and more power consumption.

\begin{table}[!h]
\begin{center}
\begin{tabular}{|l|c|c|c|c|}
\hline
Model & Params & Multi-Adds & Top1\\
\hline\hline
MobileNetV2(1.0) & 2.4M & 89M & 74.46 \\
IGCV3(0.5D)  & 1.0M & 30M & 75.75 \\
Seesaw-shareNet(v1 0.5D)  & 1.1M & 34M & \textbf {76.14} \\
\hline
\end{tabular}
\end{center}
\caption{Comparison of classification Top-1 accuracy (\%) on the CIFAR-100 dataset. exp stands for the expansion ratio within each build block.(The expansion ratio in each block is 3) }
\end{table}

\section{Conclusions and future work}

We described a very simple but neglected gourp convolution pattern that allowed us to build a family of highly efficient mobile models. Our basic building unit-Seesaw block, along with two variants Seesaw-shuffle block and Seesaw-share block, provide a novel search space for designing basic blocks to obtain better trade-off between representation capability and computational cost: Seesawnet unit introduces uneven group convolution for basic block design, Seesaw-shuffle block and Seesaw-share block can enable cross-group information flow for uneven group convolutions with and without channel permute/shuffle operations which will allows very memory-efficient inference and relies utilize standard operations present in all neural frameworks.
Due to the current optimization status of deep learning frameworks, we found that fact that large performance gap may occur even if on same target platform, when only tiny environmental modifications were made.
However, we insist the idea that neural architecture design should not only limited certain platform, since better hardware-software co-design and fine-grained optimization for specific platform in near future may help to provide better solutions.

Refer to the performance of image classification task on cifar10/cifar100/Imagenet datasets, we expect that SeesawNet with uneven group convolution will work on other classical neural blocks on alter target tasks. Exploring this is an important direction for future research.
 
\paragraph{Acknowledgments} We would like to thank Ke Sun and Mingjie Li for their source code as well as the helpful feedback.

\section{References}

[1] 2018. Microsoft Cognitive Toolkit (CNTK). https://github.com/Microsoft/CNTK.

[2] 2018. TensorFlow. https://www.tensorflow.org/.

[3] M. Abadi, A. Agarwal, P. Barham, E. Brevdo, Z. Chen, C. Citro, G. S. Corrado, A. Davis, J. Dean, M. Devin, et al. Tensorflow: Large-scale machine learning on heterogeneous systems, 2015. Software available from tensorflow. org, 1, 2015.

[4] Chollet, F. et al. Keras. https://keras.io, 2015.

[5] Yangqing Jia, Evan Shelhamer, Jeff Donahue, Sergey Karayev, Jonathan Long, Ross B. Girshick, Sergio Guadarrama, and Trevor Darrell. 2014. Caffe: Convolutional Architecture for Fast Feature Embedding. In Proceedings of the ACM International Conference on Multimedia, MM ’14, Orlando, FL, USA, November 03 - 07, 2014. 675–678.

[6] 2018. Caffe2 deep learning framework. https://github.com/caffe2/caffe2.

[7] 2018. pytorch. http://pytorch.org/.

[8] Tianqi Chen, Mu Li, Yutian Li, Min Lin, Naiyan Wang, Minjie Wang, Tianjun Xiao, Bing Xu, Chiyuan Zhang, and Zheng Zhang. 2015. MXNet: A Flexible and Efficient Machine Learning Library for Heterogeneous Distributed Systems. CoRR abs/1512.01274 (2015).

[9] 2018. TensorFlow Lite. https://www.tensorflow.org/mobile/tflite/.

[10] 2018. Tencent ncnn deep learning framework. https://github.com/Tencent/ncnn.

[11] 2018. Mobile AI Compute Engine. https://github.com/XiaoMi/mace.

[12] A. Krizhevsky, I. Sutskever, and G. E. Hinton. Imagenet classification with deep convolutional neural networks. In
Advances in neural information processing systems, pages 1097–1105, 2012.

[13] Simonyan, K. and Zisserman, A., 2014. Very deep convolutional networks for large-scale image recognition. arXiv preprint arXiv:1409.1556.

[14] Szegedy, C., Liu, W., Jia, Y., Sermanet, P., Reed, S., Anguelov, D., Erhan, D., Vanhoucke, V. and Rabinovich, A., 2015. Going deeper with convolutions. In Proceedings of the IEEE conference on computer vision and pattern recognition (pp. 1-9).

[15] K. He, X. Zhang, S. Ren, and J. Sun. Deep residual learning for image recognition. arXiv preprint arXiv:1512.03385, 2015.

[16] He, K., Zhang, X., Ren, S. and Sun, J., 2016, October. Identity mappings in deep residual networks. In European conference on computer vision (pp. 630-645). Springer, Cham.

[17] F. N. Iandola, M. W. Moskewicz, K. Ashraf, S. Han, W. J. Dally, and K. Keutzer. Squeezenet: Alexnet-level accuracy
with 50x fewer parameters and¡ 1mb model size. arXiv preprint arXiv:1602.07360, 2016.

[18] F. Chollet. Xception: Deep learning with depthwise separable convolutions. arXiv preprint arXiv:1610.02357v2, 2016.

[19] C. Szegedy, S. Ioffe, and V. Vanhoucke. Inception-v4, inception-resnet and the impact of residual connections on
learning. arXiv preprint arXiv:1602.07261, 2016.

[20] Howard, A. G.; Zhu, M.; Chen, B.; Kalenichenko, D.;Wang, W.; Weyand, T.; Andreetto, M.; and Adam, H. 2017. Mobilenets:
Efficient convolutional neural networks for mobile vision applications. arXiv preprint arXiv:1704.04861.

[21] Zhang, X., Zhou, X., Lin, M., Sun, J.: Shufflenet: An extremely effcient convolutional neural network for mobile devices. arXiv preprint arXiv:1707.01083 (2017)

[22] Iandola, F. N.; Han, S.; Moskewicz, M. W.; Ashraf, K.; Dally, W. J.; and Keutzer, K. 2016. Squeezenet: Alexnet level
accuracy with 50x fewer parameters and¡ 0.5 mb model size. arXiv preprint arXiv:1602.07360.

[23] Sandler, M., Howard, A., Zhu, M., Zhmoginov, A., Chen, L.C.: Inverted residuals and linear bottlenecks: Mobile networks for classification, detection and segmentation. arXiv preprint arXiv:1801.04381 (2018).

[24] Ma, Ningning, Xiangyu Zhang, Hai-Tao Zheng and Jian Sun. “ShuffleNet V2: Practical Guidelines for Efficient CNN Architecture Design.” ECCV (2018).

[25] Zhang, T., Qi, G.J., Xiao, B., Wang, J.: Interleaved group convolutions for deep neural networks. In: International Conference on Computer Vision. (2017)

[26] Guotian Xie, Jingdong wang, Ting Zhang, Jianhuang Lai, Richang Hong, and GuoJun Qi. Igcv2: Interleaved structured sparse convolutional neural networks. In CVPR, 2018.

[27] Sun, K., Li, M., Liu, D., Wang, J.: Igcv3: Interleaved low-rank group convolutions for efficient deep neural networks. arXiv preprint arXiv:1806.00178 (2018)

[28] Deng, J., Dong, W., Socher, R., Li, L.J., Li, K., Fei-Fei, L.: Imagenet: A large-scale hierarchical image database. In: Computer Vision and Pattern Recognition, 2009. CVPR 2009. IEEE Conference on, IEEE (2009) 248{255

[29] Russakovsky, O., Deng, J., Su, H., Krause, J., Satheesh, S., Ma, S., Huang, Z., Karpathy, A., Khosla, A., Bernstein, M., et al.: Imagenet large scale visual recognition challenge. International Journal of Computer Vision 115(3) (2015) 211{252

[30] Lin, T.Y., Maire, M., Belongie, S., Hays, J., Perona, P., Ramanan, D., Dollár, P. and Zitnick, C.L., 2014, September. Microsoft coco: Common objects in context. In European conference on computer vision (pp. 740-755). Springer, Cham.

[31] Alex Krizhevsky. Learning multiple layers of features from tiny images. Technical report, 2009.

[32] Antonio Torralba, Robert Fergus, and William T. Freeman. 80 million tiny images: A large data set for nonparametric object and scene recognition. IEEE Trans. Pattern Anal. Mach. Intell., 30(11):1958–1970, 2008. doi: 10.1109/TPAMI.2008.128.

[33] Huang, G., Liu, Z., Weinberger, K.Q., van der Maaten, L.: Densely connected convolutional networks. In: Proceedings of the IEEE conference on computer vision and pattern recognition. Volume 1. (2017)

[34] Huang, G., Liu, S., van der Maaten, L., Weinberger, K.Q.: Condensenet: An efficient densenet using learned group convolutions. arXiv preprint arXiv:1711.09224 (2017)

[35] He, K., Girshick, R. and Dollár, P., 2018. Rethinking imagenet pre-training. arXiv preprint arXiv:1811.08883.

[36] Ignatov, A., Timofte, R., Chou, W., Wang, K., Wu, M., Hartley, T. and Van Gool, L., 2018. Ai benchmark: Running deep neural networks on android smartphones. In Proceedings of the European Conference on Computer Vision (ECCV) (pp. 0-0).

[37] Han, S., Mao, H. and Dally, W.J., 2015. Deep compression: Compressing deep neural networks with pruning, trained quantization and huffman coding. arXiv preprint arXiv:1510.00149.

[38] Jacob, B., Kligys, S., Chen, B., Zhu, M., Tang, M., Howard, A., Adam, H. and Kalenichenko, D., 2018. Quantization and training of neural networks for efficient integer-arithmetic-only inference. In Proceedings of the IEEE Conference on Computer Vision and Pattern Recognition (pp. 2704-2713).

[39] Gordon, A., Eban, E., Nachum, O., Chen, B., Wu, H., Yang, T.J. and Choi, E., 2018. Morphnet: Fast and simple resource-constrained structure learning of deep networks. In Proceedings of the IEEE Conference on Computer Vision and Pattern Recognition (pp. 1586-1595).

[40] Yang, T.J., Howard, A., Chen, B., Zhang, X., Go, A., Sandler, M., Sze, V. and Adam, H., 2018. Netadapt: Platform-aware neural network adaptation for mobile applications. In Proceedings of the European Conference on Computer Vision (ECCV) (pp. 285-300).

[41] Elsken, T., Metzen, J.H. and Hutter, F., 2018. Neural architecture search: A survey. arXiv preprint arXiv:1808.05377.

[42] Tan, M., Chen, B., Pang, R., Vasudevan, V., Sandler, M., Howard, A., and Le, Q. V. (2019). Mnasnet: Platform-aware neural architecture search for mobile. In Proceedings of the IEEE Conference on Computer Vision and Pattern Recognition (pp. 2820-2828).

[43] He K, Girshick R, Dollár P. Rethinking ImageNet Pre-training[J]. 2018.

\end{document}